\documentclass[letterpaper]{article} % DO NOT CHANGE THIS
\usepackage{aaai25}  % DO NOT CHANGE THIS
\usepackage{times}  % DO NOT CHANGE THIS
\usepackage{helvet}  % DO NOT CHANGE THIS
\usepackage{courier}  % DO NOT CHANGE THIS
\usepackage[hyphens]{url}  % DO NOT CHANGE THIS
\usepackage{graphicx} % DO NOT CHANGE THIS
\usepackage{natbib}  % DO NOT CHANGE THIS AND DO NOT ADD ANY OPTIONS TO IT
\usepackage{caption} % DO NOT CHANGE THIS AND DO NOT ADD ANY OPTIONS TO IT
\usepackage{algorithm}
\usepackage{algorithmic}

\usepackage{algorithm}
\usepackage{algorithmic}
\usepackage{amssymb}
\usepackage{amsmath} 
\usepackage{pifont}
\usepackage{booktabs}
\usepackage{multirow}

\usepackage{makecell}

\def\figureautorefname{Fig.}%
\def\tableautorefname{Table }%

\newcommand{\valstd}[2]{$#1 {\scriptstyle \,\pm\, #2}$}
\newcommand{\valstdb}[2]{$\mathbf{#1} {\scriptstyle \,\pm\, #2}$}
\newcommand{\valstdu}[2]{$\underline{#1} {\scriptstyle \,\pm\, #2}$}

\usepackage{newfloat}
\usepackage{listings}
\DeclareCaptionStyle{ruled}{labelfont=normalfont,labelsep=colon,strut=off} % DO NOT CHANGE THIS
\floatstyle{ruled}
\newfloat{listing}{tb}{lst}{}
\floatname{listing}{Listing}
\title{TimeCHEAT: A Channel Harmony Strategy for Irregularly Sampled Multivariate Time Series Analysis}
\author{
    Jiexi Liu\textsuperscript{\rm 1,2},
    Meng Cao\textsuperscript{\rm 1,2},
    Songcan Chen\textsuperscript{\rm 1,2}\thanks{Corresponding Author}
}
\affiliations{
    \textsuperscript{\rm 1}College of Computer Science and Technology, Nanjing University of Aeronautics and Astronautics\\
    \textsuperscript{\rm 2}MIIT Key Laboratory of Pattern Analysis and Machine Intelligence\\    
    \{liujiexi, meng.cao, s.chen\}@nuaa.edu.cn
}

\usepackage{bibentry}
\begin{document}

\maketitle
\begin{abstract}
Irregularly sampled multivariate time series (ISMTS) are prevalent in reality. Due to their non-uniform intervals between successive observations and varying sampling rates among series, the channel-independent (CI) strategy, which has been demonstrated more desirable for \textit{complete multivariate time series forecasting} in recent studies, has failed. This failure can be further attributed to the sampling sparsity, which provides insufficient information for effective CI learning, thereby reducing its capacity.  When we resort to the channel-dependent (CD) strategy, even higher capacity cannot mitigate the potential loss of diversity in learning similar embedding patterns across different channels. We find that existing work considers CI and CD strategies to be mutually exclusive, primarily because they apply these strategies to the global channel. However, we hold the view that \textit{channel strategies do not necessarily have to be used globally}. Instead, by appropriately applying them locally and globally, we can create an opportunity to take full advantage of both strategies.  This leads us to introduce the Channel Harmony ISMTS Transformer (TimeCHEAT), which \textit{utilizes the CD strategy locally and the CI strategy globally}. Specifically, we segment the ISMTS into sub-series level patches. Locally, the CD strategy aggregates information within each patch for time embedding learning, maximizing the use of relevant observations while reducing long-range irrelevant interference. Here, we enhance generality by transforming embedding learning into an edge weight prediction task using bipartite graphs, eliminating the need for special prior knowledge. Globally, the CI strategy is applied across patches, allowing the Transformer to learn individualized attention patterns for each channel. Experimental results indicate our proposed TimeCHEAT demonstrates competitive state-of-the-art performance across three mainstream tasks including classification, forecasting and interpolation.
\end{abstract}

\begin{figure}[t]
\centerline{\includegraphics[width=0.95\linewidth]{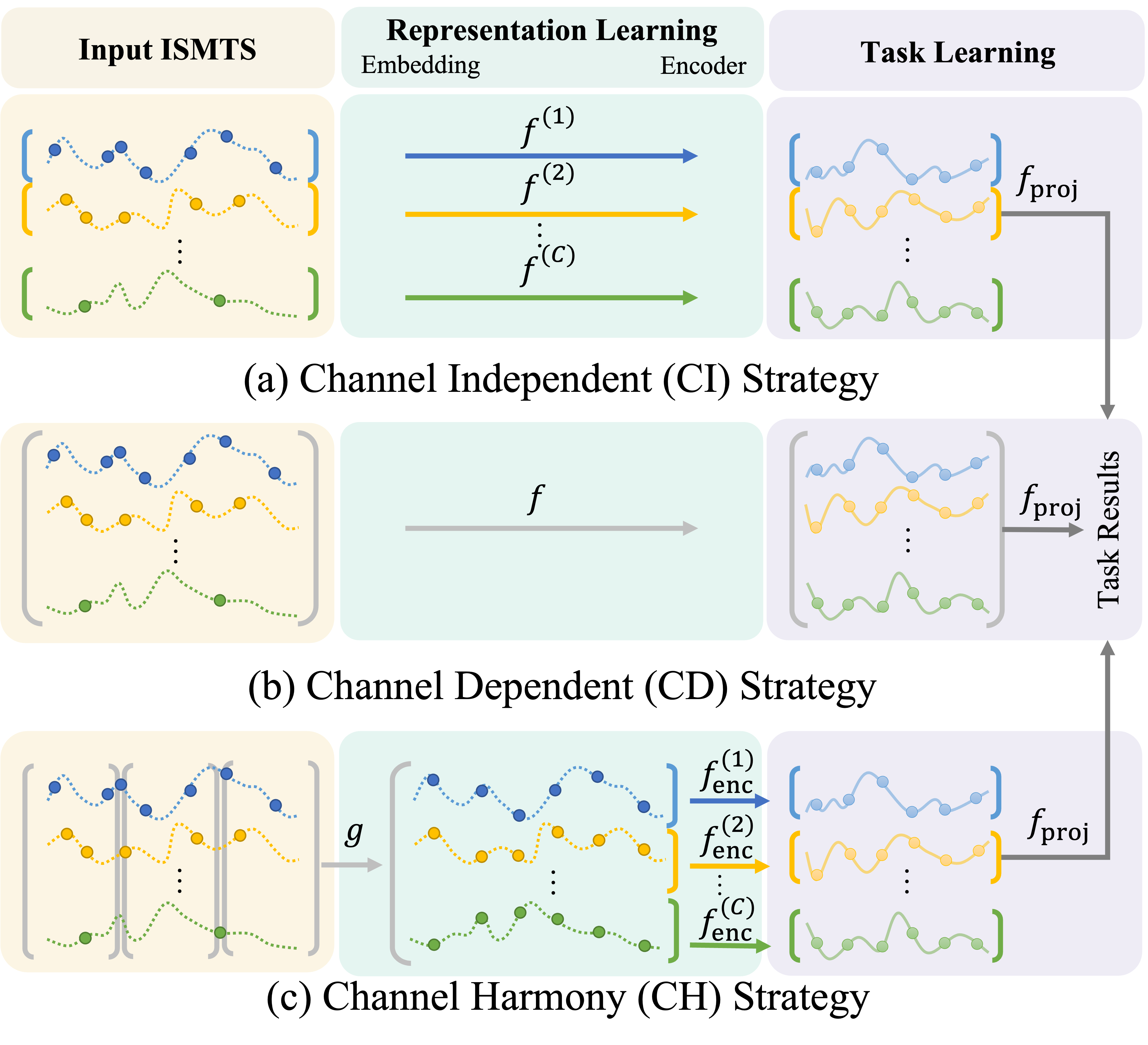}}
\caption{The difference between the 3 kinds of channel strategies.}
\label{fig:CICD}
\end{figure}

\section{Introduction}
\label{sec:introduction}
Irregularly sampled multivariate time series (ISMTS) are ubiquitous in real-world scenarios, such as healthcare \cite{goldberger2000physiobank,reyna2020early}, meteorology \cite{schulz1997spectrum,cao2018brits} and transportation \cite{chen2022nonstationary, tang2020joint}. Due to sensor malfunctions, transmission distortions, cost-reduction strategies, etc, ISMTS are characterized by inconsistent intervals between consecutive data points within a channel, asynchronous sampling across multiple channels and sometimes sampling sparsity. Such characteristics, arising from irregularities and multi-channels, pose a significant challenge to classical machine learning methods, which require the data to be defined in a consistent, fixed-dimensional feature space with constant intervals between successive timestamps.

Recent studies have sought to address this challenge using various approaches. One common method is a two-step process that treats ISMTS as synchronized, regularly sampled Normal Multivariate Time Series (NMTS) data with missing values, focusing on imputation strategies \cite{che2018recurrent, yoon2018gain, camino2019improving, tashiro2021csdi, le2021sa, chen2022nonstationary, fan2022dynamic, du2023saits}. However, accurate imputation is challenging, therefore separating imputation from downstream tasks may distort the underlying relationships and introduce substantial noise, leading to suboptimal results \cite{zhang2021graph, wu2021dynamic, agarwal2023modelling, sun2024time}. Since imputation needs to fully utilize information under the ISMTS scenario, two-step methods often use the channel-dependent (CD) strategy, merging all input dimensions in the learning process. While end-to-end models, which have gained considerable attention recently, have demonstrated superior performance over two-step approaches \cite{le2021sa}. Some of these models treat ISMTS as discrete-time series, aggregating sample points from individual or multiple channels to create unified features \cite{wu2021dynamic,agarwal2023modelling}, while others preserve the continuous temporal dynamics of ISMTS data, often processing each channel independently \cite{de2019gru,kidger2020neural,schirmer2022modeling,jhin2022exit,chowdhury2023primenet}. The end-to-end methods utilize CD and channel-independent (CI) strategies arbitrarily based on the existing models or experimental results. Therefore, despite the numerous time series models proposed to tackle irregular sampling in ISMTS, the challenge of effectively managing channel interactions still remains unresolved. 

Since the interactions between multiple channels in ISMTS data are complex, a deeper understanding of these interactions lead to more accurate and insightful analysis. Given the advantages of end-to-end methods above, this paper focuses on these approaches. As shown in \figureautorefname \ref{fig:CICD}, we divide the ISMTS analysis process into three main steps: ISMTS input, representation learning, and downstream tasks. Our primary focus is on representation learning, which includes both embedding learning and encoding. We do not separate these in \figureautorefname \ref{fig:CICD} (a) and (b) for two existing primary channel strategies CI and CD because the strategy remains consistent unchanged throughout the process. The \textbf{CI strategy} as in \figureautorefname \ref{fig:CICD}(a), which uses individual models for each channel, works well in NMTS forecasting \cite{nie2023a}. In ISMTS analysis, CI allows channels to be processed according to their unique sampling patterns without forcing synchronization, preserving data integrity. However, CI struggles with limited generalizability and robustness on unseen channels \cite{han2024capacity} and the varying sampling rates across channels can result in the loss of crucial context, and channels with sparse sampling may fail to provide sufficient information for effective learning. On the other hand, the \textbf{CD strategy} in \figureautorefname{\ref{fig:CICD}}(b) models all channels together, capturing complex temporal patterns but risks oversmoothing and struggles to fit individual channels, particularly when channel similarity is low. 
 
To effectively manage channel interactions, this paper aims to balance necessary individual intra-channel treatment with inter-channel dependencies simultaneously.  While existing work often treats CI and CD strategies as mutually exclusive, applying them only at a global level, we hold the view that these strategies can be more effective when used both locally and globally. This approach create an opportunity to fully leverage the advantages of both strategies. Accordingly, we propose the \textbf{C}hannel \textbf{H}armony Irr\textbf{E}gularly S\textbf{A}mpled Multivariate Time Series \textbf{T}ransformer (\textbf{TimeCHEAT}), which applies the CD strategy locally and the CI strategy globally, as illustrated in \figureautorefname \ref{fig:CICD}(c). By applying the CD strategy locally and the CI strategy globally, the model achieves an effective balance between capturing detailed, context-specific information and preserving broader, channel-specific patterns. This hybrid approach enables more accurate and insightful analysis of ISMTS data, addressing the complex interactions between channels without sacrificing either detail or generality.

Specifically, we begin by dividing the ISMTS into subseries-level patches. Within each patch, we apply the CD strategy locally to effectively learn time embeddings. This approach aggregates nearby observations and neighborhood channel information, leveraging local smoothness to highlight their importance while avoiding interference from distant, irrelevant data. Traditional methods for time embedding learning often assume that larger time intervals weaken dependencies and relationships, leading to strong inductive biases. This can cause key timestamps with independent positions to be overlooked, thus affecting the extraction of important patterns. To address this, we transform the embedding learning process into an edge weight prediction task using bipartite graphs, which provides a more straightforward and generalizable learning method without above assumption. Additionally, we apply the CI strategy globally across patches, using the Transformer as the backbone encoder to capture individual attention patterns for each channel.

Our main contributions can be summarized as follows:
\begin{itemize}
    \item We are the first to explore channel strategies for ISMTS analysis, proposing a novel and unified approach that better balances individual channel treatment with cross-channel modeling.
    \item We design a special time embedding method that can directly learn fix-length time embedding for ISMTS data without introducing special inductive bias.
    \item We are not limited to a specific forecasting task but attempt to propose a task-general channel strategy for ISMTS data, including classification, forecasting and interpolation. 
\end{itemize}

\section{Related Work}
\label{sec:relatdwork}
\subsection{Irregularly Sampled Multivariate Time Series Analysis Models}

An effective approach to analyzing ISMTS relies on understanding their unique properties. One natural idea is the two-step method, which treats ISMTS as NMTS with missing values and imputes the missingness before performing downstream tasks \cite{che2018recurrent,yoon2018gain,camino2019improving,tashiro2021csdi,chen2022nonstationary,fan2022dynamic,du2023saits,wang2024deep}. However, most two-step methods may distort the underlying relationships, introducing unsuitable assumptions and substantial noise due to incorrect imputation \cite{zhang2021graph,wu2021dynamic,agarwal2023modelling}, ultimately compromising the accuracy of downstream tasks. Therefore, recent work has shifted towards using end-to-end methods, which have been shown to outperform two-step methods both experimentally and theoretically \cite{le2021sa}. One approach treats ISMTS as time series with discrete timestamps, focusing on handling the irregularities by aggregating all sample points of either a single channel or all channels to extract a unified feature  \cite{wu2021dynamic,agarwal2023modelling}. Others leverage the inherent continuity of time, thereby preserving the ongoing temporal dynamics present in ISMTS data \cite{de2019gru,rubanova2019latent,kidger2020neural,schirmer2022modeling,jhin2022exit,chowdhury2023primenet}. 

Due to these advancements, the issue of irregular sampling has been initially addressed. However, the relationships between channels have yet to be discussed. Since the channel strategies for most ISMTS analysis models simply merge all channels as input without considering the special inter-channel correlation nor treating different channels individually.

\begin{figure*}[t]
\centerline{\includegraphics[width=0.9\linewidth]{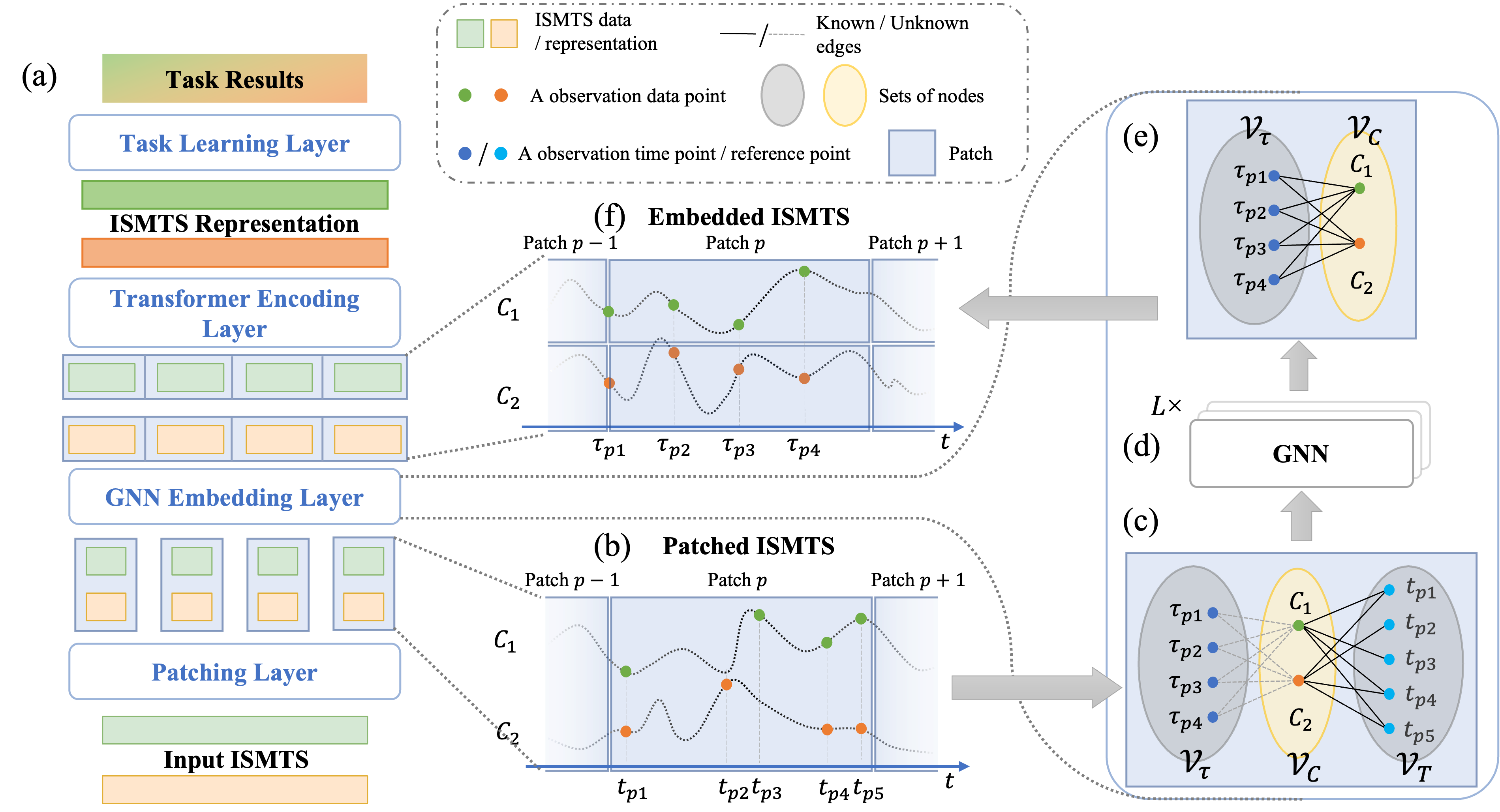}}
\caption{Overview of TimeCHEAT framework, shown in (a), containing $3$ main steps, including ISMTS \textbf{embedding learning}, \textbf{Transformer encoding} and \textbf{task learning}.
(b) is the patched ISMTS data with $p$ patches. (c) is the initially established bipartite graph with known edges between channel and observation time step nodes and unknown edges between channel and reference point modes. (d) GNN module for unknown edges learning. (e) Learned edges between channel and reference point nodes. (f) Graph to embedding process to produce embedded ISMTS. }
\label{framework}
\end{figure*}

\subsection{Channel Strategies for Time Series Analysis}
Channel Strategies exist in all deep learning multivariate time series learning models, but the discussion of this topic mainly focuses on the \textit{complete multivariate} time series \textit{forecasting} task. The essential challenge of channel strategies for other downstream tasks as well as for ISMTS data from the model design perspective has not been solved. Before the discovery in \citet{montero2021principles} and the advent of PatchTST \cite{nie2023a}, the CD strategy was the mainstream approach in time series deep learning models \cite{wu2021autoformer, liu2021pyraformer, zhou2022fedformer} that tried to fully use information across channels. Conversely, the emergence of PatchTST breaks new ground in CI strategy, followed by several inspiring work \cite{li2023revisiting, zeng2023transformers, chen2024similarity, han2024capacity} and conclude that CI has high capacity and low robustness, whereas CD is the opposite. Since existing work holds the view that CI and CD strategies are mutually exclusive, existing works focus on improving one of the strategies. \citet{han2024capacity} design Predict Residuals with Regularization to incorporate a regularization term in the objective to address the non-robustness of the CD strategy and encourage smoothness in future forecasting. Moreover, \citet{chen2024similarity} dynamically group channels characterized by intrinsic similarities and leveraging cluster identity instead of channel identity. A similar idea has been proposed for ISMTS in LIFE \cite{zhang2021life} which collects credible and correlated channels to build individual features. Therefore, it remains challenging to develop a balanced channel strategy that contains advantages from both strategies. More importantly, the potential and effect of channel strategy for ISMTS remain under-explored.

\section{Preliminaries}
\label{sec:preliminaries}

\paragraph{Notations.}

A set of ISMTS is a finite sequence of tuples $S = (X, M, Y) \in \mathbb{R}^d \times \{0, 1\}^d \times \mathbb{R}^{d'}$, where $X$ is the whole time series, $M$ is a missingness indicator, and $Y$ is a response of interest. Here, $d = N \times C \times T $ shows the dimensions of the input ISMTS, where $N$ is the number of instances, $C$ denotes the number of variates (i.e., channels) and $T$ is the length of the time series. While $d' = 1 $ for the classification task and $d' = N \times C \times T'$ for the forecasting problem, in which $T'$ indicates the forecasting horizon. For each realization, $(x, m, y)$, $m_{c, t} = 0$ indicates there is no observation at timestamp $t$ in channel $c$, and $m_{c, t} = 1$ means it is observed. \textit{ Here, for brevity, we omit the data case index $n$ for the $n$-th instance when the context is clear. }

\paragraph{Channel Dependent (CD) and Channel Independent (CI).}
We define multivariate time series in accordance with previous literature \cite{han2024capacity, murtagh2012multivariate, zhou2021informer}, which consists of two or more real-valued channels that depend on time. Other sequential data with extra dimensions like spatio-temporal data \cite{tan2023openstl} or discrete values like natural language \cite{kenton2019bert} are not included in this paper.

While different from existing work, our TimeCHEAT focuses on learning a set of nonlinear embedding functions $F$ that can map the input ISMTS into best-described representations for further analysis. Under this scenario, the CI strategy models each channel separately and ignores the potential channel interactions. This approach needs $C$ individual functions and each is typically denoted as $f^{(c)}: \mathbb{R}^T \rightarrow \mathbb{R}^T$, for $c = 1, \ldots, C$, where $f^{(c)}$ gets access to the ISMTS data specific to the $c$-th channel. On the contrary, the CD strategy models all the channels as a whole, using a single function $f: \mathbb{R}^{T \times C} \rightarrow \mathbb{R}^{T \times C}$. The learned representations $H \in \mathbb{R}^{T\times C}$ are utilized for further downstream tasks, including classification and forecasting in our paper.

\section{Proposed TimeCHEAT Framework}
\label{sec:timecheat}

In this paper, we propose an ISMTS analysis method that employs a novel channel harmony strategy for enhanced feature learning. As illustrated in \figureautorefname 2, the effectiveness of TimeCHEAT is largely ensured by 
1) CD strategy for ISMTS embedding ensures comprehensive integration of multivariate data within a patch.
2) Bipartite graph modeling facilitates effective embedding learning by capturing complex relationships.
3) CI strategy for representation learning allows for tailored attention to individual channel dynamics. 
We will then discuss the above key points in the following subsections.

\subsection{Locally Independent for Embedding Learning}
Embedding learning is a crucial operation in end-to-end ISMTS learning. Some existing methods assume that large time intervals weaken dependencies \cite{che2018recurrent, shukla2018interpolation, shukla2021multitime, shukla2022heteroscedastic}. This assumption can lead to unrelated points being perceived as highly related while key points are overlooked, ultimately affecting the extraction of important patterns. While others rely on Neural Ordinary Differential Equations (NeuralODEs) \cite{chen2018neural, jin2022multivariate, scholz2022latent} which are known to be slow and often require additional features to handle irregularities.

To address these challenges,
we observe that predicting edge weights in bipartite graphs can facilitate learning values based on query input, serving as an effective method for continuous-time embedding learning by the power of graph structures. It elegantly avoids the aforementioned assumptions and has been proven effective in ISMTS forecasting  \cite{grafiti2024Yalavarthi} and imputing missing data \cite{you2020handling}.

\paragraph{Graph Learning.} We begin by segmenting the ISMTS into $P$ equal-length, non-overlapping patches and constructing a bipartite graph for each patch $p$, as illustrated in \figureautorefname \ref{framework} (b) and (c). For simplicity, we omit the patch index $p$ in this subsection, as all calculations are performed within a single patch. To effectively learn the local representation of the input ISMTS, each patch can be transferred into a bipartite graph $\mathcal{G} = (\mathcal{V}, \mathcal{E})$, consisting of two disjoint node sets and an edge set: the channel node set $\mathcal{V}_C$, the timestamp node set $\mathcal{V}_T$ and the observation measurement set $\mathcal{E} \subseteq \mathcal{V}_C \times \mathcal{V}_T$. Our goal is to learn from these edges to generalize relationships across channels and arbitrary timestamps.

To generate a fixed-dimensional embedding for each patch, making it suitable for subsequent neural network processing, we predefined a set of reference points $\boldsymbol{\tau} = [\tau_1,\cdots,\tau_K]$. These reference points are regular timestamps without observations, serving as a special subset of query nodes $\mathcal{V}_\tau$. The initial connection between reference points nodes set $\mathcal{V}_\tau$ and channels $\mathcal{V}_C$ are all $0$, while the observation timestamp nodes set $\mathcal{V}_T$ and channels $\mathcal{V}_C$ have real observation value $x_{c,t}$. Here, we introduce an enhanced edge with an indicator $i$, denoted as
\begin{equation}
    E(e, i) =\left\{\begin{array}{lll}
\left(x_{c, t}, 1\right), & e \in \mathcal{E}_T \\
(0, 0),   & e \in \mathcal{E}_\tau
\end{array} \right.
\end{equation}

\noindent when an edge is observed, $i$ is set to 1, otherwise, 0.

We then introduce the Irregularity to Regularity Graph (I2RGraph), as shown in \figureautorefname \ref{framework} (c) to (e), aiming to transform irregularities into regularities by learning the edges $\mathcal{E}_\tau \subseteq \mathcal{V}_C \times \mathcal{V}_\tau$. This process can be formally expressed as follows:
\begin{equation}
    \mathcal{E}_{{\tau}} := \textnormal{I2RGraph}\left(\mathcal{V}_C, \mathcal{V}_T, \mathcal{E}_T, \mathcal{V}_{{\tau}}\right)
\end{equation}

\paragraph{Patch Embedding Learning.} When applying the CD strategy, it is insufficient to rely solely on channel ID numbers for learning. To address this, we encode the channel IDs to capture the underlying correlations between channels:
\begin{equation}
    h^{\text{node}, 0}_c = \textbf{FFN}\left(\textbf{CM}\left(c\right)\right), c \in \mathcal{V}_C
\end{equation}
Here, $\textbf{CM}$ is a learnable matrix, initially set as an identity matrix, and $c$ represents the channel ID used for encoding. $\textbf{FFN}$ denotes a fully connected layer. Notably, if $\textbf{CM}$ is not learnable, $h^{\text{node}, 0}_c$ cannot capture these underlying correlations due to the independence of one-hot encodings.

 According to \cite{shukla2021multitime}, the timestamps can be encoded as a learned sinusoidal encoding and the initial edge embedding:
\begin{equation}
    \begin{aligned}
    h_t^{\text{node}, 0} & = \sin\left(\textbf{FFN}\left(t\right)\right), t \in \mathcal{V}_T {\cup} \mathcal{V}_{\boldsymbol{\tau}}\\
    h^{\text{edge}, 0}_e & = \textbf{FFN}\left(E(e, i)\right), e \in \mathcal{E}
    \end{aligned}
\end{equation}
\noindent where ${\cup}$ denotes the union of two sets.

With initial embedding $(h^{\text{node},0})$ and $(h^{\text{edge},0})$ above, we then utilize Graph Attention Network (GNN) \cite{velivckovic2017graph, grafiti2024Yalavarthi} for further graph learning, more details are in the supplementary. The proposed GNN architecture consists of $L$ layers and the node features are updated layer wise from layer $l$ to $l + 1$ as follows:
\begin{equation} \label{eq:node:update}
        \begin{aligned}
                & h_{u}^{\text {node }, l+1} =\mathbf{MultiHead}^{(l)}\left(h_{u}^{\text {node }, l}, H_{u}, H_{u}\right)\\
                & s.t. \quad H_{u} =\left(\left[h_{v}^{\text {node }, l} \| h_{e}^{\text {edge }, l}\right]\right)_{v \in \mathcal{N}(u)}
        \end{aligned}
\end{equation}
\noindent in which $\|$ represents the concatenation operation between two vectors or matrices,  $\mathbf{MultiHead}$ is a multi-head attention block \cite{vaswani2017attention}, $\mathcal{N}(u) := \{v | e_{u,v} \in \mathcal{E}, v \in \mathcal{V} \}$ indicates the neighborhood of $u$ including all connected nodes of $u$ though edges in $\mathcal{E}$. Eq. \eqref{eq:node:update} means to search for the most relevant nodes among timestamp (channel) nodes to update the feature of a given channel (timestamp) node $u$.

The edge features are related to the corresponding channels, timestamps and edges themselves. Therefore, we adopt the following formula to update the edge features:
\begin{equation}
   \begin{aligned}
    & h_{e}^{\text {edge}, l+1} = \\
    & \quad \alpha\left(h_{e}^{\text {edge }, l} +\mathbf{FFN}^{(l)}\left( \left[ h_{c}^{\text {node }, l}\left\|h_{t}^{\text {node }, l}\right\| h_{e}^{\text {edge }, l}\right] 
    \right)\right)
    \end{aligned}
\end{equation}
\noindent in which, $\alpha$ is a non-linear activation, and a residual structure \cite{he2016deep} has been adopted for stable learning.

Finally, the output embedding of the $p$-th patch is 
\begin{equation}
    H_p = \text{I2RGraph}(h^{{\text{node}, L}}, h^{\text{edge}, L})
\end{equation}
and we will use the overall embedding $H = \left\{H_p\right\}_{p=1}^{P}$ for further encoding.

\subsection{Globally Independent for Transformer Encoding}
 After the embedding learning phase, we obtain several fixed-length patch embeddings $H$ for the ISMTS, which can be directly fed into the encoder. In the case of symbol abuse, we denote the dimension of $H$ as $P \times C \times T_P$. Subsequently, we apply a CI Transformer encoder to the patch time series, which maps the embedding  $H$  into a representation  $R$  for various downstream tasks. In other words, the subseries-level patches created during the embedding phase act as input tokens for the Transformer. This CI process, within each channel, contains a univariate time series that shares the same Transformer weights across all channels. To describe the temporal relationship, we provide the position encoding supported by the patch ID $p$ and the position $i$ for each channel as follows:
\begin{equation}
    \left\{\begin{aligned}
        PE_{p, 2i} = &  \sin\left( \frac{p}{10000^{2i/T_P}} \right)\\
         PE_{p, 2i + 1} =& \cos\left( \frac{p}{10000^{2i/T_P}} \right) 
    \end{aligned}\right.
    \label{TEmbedding}
\end{equation}

To obtain the final representation $R$, we then feed the embedding $H$ into a multi-head self-attention module with three learnable variable matrices, which are $K$, $Q$, and $V$.
 \begin{equation}
     R_c = \textbf{MultiHead}\left(KH_c^{PE}, QH_c^{PE}, VH_c^{PE}\right)
 \end{equation}
\noindent where $R_c$ is the $c$-th channel of $R$, $H_c^{PE} = H_c + PE$, and $H_c \in \mathbb{R}^{P \times T_P}$ denotes $c$-th channel of $H$.

\begin{table*}[!t]
% \scriptsize
\centering
% \small
\scriptsize
\resizebox{0.788\textwidth}{!}{ 
\begin{tabular}{l|ll|ll|llll}
\toprule
& \multicolumn{2}{c|}{P19} & \multicolumn{2}{c|}{P12} & \multicolumn{4}{c}{PAM} \\ \cmidrule{2-9}
\multirow{-2}{*}{Methods} & AUROC & AUPRC & AUROC & AUPRC & Accuracy & Precision & Recall & F1 score \\ \midrule
Transformer & \valstd{80.7}{3.8} & \valstd{42.7}{7.7} & \valstd{83.3}{0.7} & \valstd{47.9}{3.6} & \valstd{83.5}{1.5} & \valstd{84.8}{1.5} & \valstd{86.0}{1.2} & \valstd{85.0}{1.3} \\
Trans-mean & \valstd{83.7}{1.8} & \valstd{45.8}{3.2} & \valstd{82.6}{2.0} & \valstd{46.3}{4.0} &	\valstd{83.7}{2.3} &\valstd{84.9}{2.6} & \valstd{86.4}{2.1}&\valstd{85.1}{2.4}\\
GRU-D & \valstd{83.9}{1.7} & \valstd{46.9}{2.1} & \valstd{81.9}{2.1} & \valstd{46.1}{4.7} & \valstd{83.3}{1.6} & \valstd{84.6}{1.2} & \valstd{85.2}{1.6} & \valstd{84.8}{1.2}\\
SeFT & \valstd{81.2}{2.3} & \valstd{41.9}{3.1} & \valstd{73.9}{2.5} & \valstd{31.1}{4.1} & \valstd{67.1}{2.2} & \valstd{70.0}{2.4} & \valstd{68.2}{1.5} & \valstd{68.5}{1.8}\\
mTAND & \valstd{84.4}{1.3} & \valstd{50.6}{2.0} & \valstd{84.2}{0.8} & \valstd{48.2}{3.4} & \valstd{74.6}{4.3} & \valstd{74.3}{4.0} & \valstd{79.5}{2.8} & \valstd{76.8}{3.4}\\ 
IP-Net & \valstd{84.6}{1.3} & \valstd{38.1}{3.7} & \valstd{82.6}{1.4} & \valstd{47.6}{3.1} & \valstd{74.3}{3.8} & \valstd{75.6}{2.1} & \valstd{77.9}{2.2} & \valstd{76.6}{2.8}\\
DGM$^2$-O & \valstd{86.7}{3.4} & \valstd{44.7}{11.7} & \valstd{84.4}{1.6} & \valstd{47.3}{3.6} & \valstd{82.4}{2.3} & \valstd{85.2}{1.2} & \valstd{83.9}{2.3} & \valstd{84.3}{1.8}\\
MTGNN & \valstd{81.9}{6.2}  & \valstd{39.9}{8.9} & \valstd{74.4}{6.7} & \valstd{35.5}{6.0} & \valstd{83.4}{1.9} & \valstd{85.2}{1.7} & \valstd{86.1}{1.9} & \valstd{85.9}{2.4} \\
Raindrop & \valstd{87.0}{2.3} & \valstd{51.8}{5.5} & \valstd{82.8}{1.7} & \valstd{44.0}{3.0} & \valstd{88.5}{1.5} & \valstd{89.9}{1.5} & \valstd{89.9}{0.6} & \valstd{89.8}{1.0}\\
ViTST &\valstdu{89.2}{2.0} &\valstdu{53.1}{3.4} &\valstdb{85.1}{0.8}  &\valstdb{51.1}{4.1} & \valstdu{95.8}{1.3} &\valstdu{96.2}{1.3} &\valstdu{96.1}{1.1} & \valstdu{96.5}{1.2}\\
\midrule
\textbf{TimeCHEAT} &\valstdb{89.5}{1.9} &\valstdb{56.1}{4.6} &\valstdu{84.5}{0.7}  &\valstdu{48.2}{1.9} & \valstdb{ 96.5}{0.6} &\valstdb{97.1}{0.5} &\valstdb{96.9}{0.6} & \valstdb{97.0}{0.5}\\
\bottomrule
\end{tabular}
}
\caption{Comparison with the baseline methods on ISMTS \textbf{classification} task. }
\label{tab:main_result}
\end{table*}

\begin{table*}[t]
\centering
%\footnotesize
% \small
\scriptsize
\tabcolsep=1.508em
\begin{tabular}[h]{c| c |c |c |c |c}
     \toprule
     {\bf Model} & \multicolumn{5}{c}{{\bf Mean Squared Error} $(\times 10^{-3})$} \\
     \midrule
     Observed \%  & $50\%$ & $60\%$ & $70\%$ & $80\%$ & $90\%$ \\
          \midrule
     RNN-VAE &  \valstd{13.418}{0.008} & \valstd{12.594}{0.004} & \valstd{11.887}{0.005} & \valstd{11.133}{0.007} & \valstd{11.470}{0.006}\\
     L-ODE-RNN & \valstd{8.132}{0.020} & \valstd{8.140}{0.018} & \valstd{8.171}{0.030} & \valstd{8.143}{0.025} & \valstd{8.402 }{0.022} \\
     L-ODE-ODE &  \valstd{6.721}{0.109} & \valstd{6.816}{0.045} & \valstd{6.798}{0.143} & \valstd{6.850}{0.066} & \valstd{7.142}{0.066}\\
     mTAND-Full & \valstdb{4.139}{0.029} & \valstdu{4.018}{0.048} & \valstdu{4.157}{0.053} &  \valstdu{4.410}{0.149} & \valstdu{4.798}{0.036} \\
     \midrule
     \textbf{TimeCHEAT} & \valstdu{4.185}{0.030} &  \valstdb{3.981}{0.016} &  \valstdb{3.657}{0.022} &   \valstdb{3.642}{0.036} &  \valstdb{3.686}{0.009} \\

    \bottomrule
\end{tabular}
\caption{Comparison with the baseline methods on ISMTS \textbf{interpolation} task on PhysioNet.} 
\label{table:phy_interp}
\end{table*}

\begin{table}[t]
	\centering
      \scriptsize
      \tabcolsep=0.15em
	\begin{tabular}{c|c|c|c|c}
		\toprule  \multicolumn{1}{c}{Methods}  & \multicolumn{1}{|c}{USHCN} & \multicolumn{1}{|c}{MIMIC-III} & \multicolumn{1}{|c}{MIMIC-IV} &\multicolumn{1}{|c}{Physionet12}\\ 
		\midrule
		DLinear+ & \valstd{0.347}{0.065} & \valstd{0.691}{0.016} &\valstd{0.557}{0.001}  &\valstd{0.380}{0.001} \\
		NLinear+ & \valstd{0.452}{0.101} & \valstd{0.726}{0.019} & \valstd{0.620}{0.002}  & \valstd{0.382}{0.001}   \\
		Informer+ & \valstd{0.320}{0.047} & \valstd{0.512}{0.064} &  \valstd{0.420}{0.007} &  \valstd{0.347}{0.001}\\
		FedFormer+ & \valstd{2.990}{0.476} & \valstd{1.100}{0.059} & \valstd{2.135}{0.304} & \valstd{0.455}{0.004}\\
		\midrule
		NeuralODE-VAE & \valstd{0.960}{0.110} & \valstd{0.890}{0.010} & $-$  & $-$\\
		GRU-Simple & \valstd{0.750}{0.120} & \valstd{0.820}{0.050} & $-$ &$-$\\
		GRU-D & \valstd{0.530}{0.060}& \valstd{0.790}{0.060} & $-$ &$-$\\
		T-LSTM & \valstd{0.590}{0.110}& \valstd{0.620}{0.050} & $-$ & $-$\\
		mTAND & \valstd{0.300}{0.038} & \valstd{0.540}{0.036} & ME & \valstd{0.315}{0.002} \\
		GRU-ODE-Bayes & \valstd{0.430}{0.070} & \valstd{0.480}{0.480} & \valstd{0.379}{0.005} & \valstd{0.329}{0.004}\\
		Neural Flow & \valstd{0.414}{0.102} & \valstd{0.490}{0.004} & \valstd{0.364}{0.008}  & \valstd{0.326}{0.004} \\
		CRU & \valstd{0.290}{0.060} & \valstd{0.592}{0.049} & ME  & \valstd{0.379}{0.003} \\
		GraFITi& \valstdu{0.272}{0.047} &  \valstdb{0.396}{0.030} & \valstdb{0.225}{0.001} & \valstdb{0.286}{0.001} \\
        \midrule 
        \textbf{TimeCHEAT} & \valstdb{0.266}{0.069} &  \valstdu{0.462}{0.034} & \valstdu{0.273}{0.002} & \valstdu{0.290}{0.001} \\
    \bottomrule
	\end{tabular}
 \caption{Experimental results for \textbf{forecasting} next three time steps. $-$ indicates no published results. ME indicates a Memory Error.}
	\label{tab:fore}
 \end{table}

\paragraph{ISMTS Analysis Tasks.}
% Here, we specifically concentrate on classification tasks as a representative example of supervised learning. The loss function is
Here, for classification, the loss function is defined as:
\begin{equation}
    \mathcal{L} = \frac{1}{N} \sum_{n=1}^{N}{ {\ell_{CE}\left(\operatorname{CLS}\left({R}_n\right), y_n\right)}}
\end{equation}
\noindent where $\operatorname{CLS}\left(\cdot\right)$ denotes the projection head for classification, and $\ell_{CE}\left(\cdot\right)$ denotes the cross-entropy loss.

While for interpolation and forecasting, the associated loss function is defined as:
\begin{equation}
    \mathcal{L} = \frac{1}{N} \sum_{n=1}^{N}{\|M'_n \odot(\operatorname{PRE}\left({R}_n\right)-y_n)\|_{2}^{2}}
\end{equation}
\noindent where $\operatorname{PRE}\left(\cdot\right)$ denotes the projection head for forecasting or imputation. As observations might be missing also in the groundtruth data, to measure the accuracy we average an element-wise loss function over only valid values using $M'_n$.

\section{Experiments}
\label{sec:experiments}
In this section, we show the effectiveness of the TimeCHEAT framework across $3$ mainstream time series analysis tasks, including classification, interpolation, and forecasting. The results are reported as mean and standard deviation values calculated over $5$ independent runs. The \textbf{bold} font highlights the top-performing method, while the \underline{underlined} text marks the runner-up. Additional experimental setup details are provided in the Appendix due to space constraints.

\subsection{Time Series Classification}

\paragraph{Datasets and Experimental Settings.}
We use real-world datasets from healthcare to human activity domains for classification tasks: (1) \textbf{P19} \cite{reyna2020early}, with a missing ratio of $94.9\%$, includes $38,803$ patients monitored by $34$ sensors. (2) \textbf{P12} \cite{goldberger2000physiobank} contains temporal measurements from $36$ sensors of $11,988$ patients during the first 48 hours of ICU stay, with a missing ratio of $88.4\%$. (3) \textbf{PAM} \cite{reiss2012introducing} consists of $5,333$ segments from $8$ daily activities, measured by $17$ sensors, with a missing ratio of $60.0\%$. \textit{P19 and P12 are \textbf{imbalanced} binary label datasets} while PAM dataset contains $8$ classes. 

Following standard practice, we randomly split each dataset into training ($80\%$), validation ($10\%$), and test ($10\%$) sets, using fixed indices across all methods. For the imbalanced P12 and P19 datasets, we evaluate performance using the area under the receiver operating characteristic curve (AUROC) and the area under the precision-recall curve (AUPRC). For the nearly balanced PAM dataset, we use Accuracy, Precision, Recall, and F1 Score. For all of the above metrics, higher values indicate better performance.

\paragraph{Main Classification Results.}
We compare TimeCHEAT with $10$ state-of-the-art methods for irregularly sampled time series classification, including Transformer \cite{vaswani2017attention}, Trans-mean, GRU-D \cite{che2018recurrent}, SeFT \cite{horn2020set}, mTAND \cite{shukla2021multitime}, IP-Net \cite{shukla2018interpolation}, DGM$^2$-O \cite{wu2021dynamic}, MTGNN \cite{wu2020connecting}, Raindrop \cite{zhang2021graph}, and ViTST \cite{li2023time}. Since \textit{mTAND has demonstrated superiority over various recurrent models} such as RNNImpute \cite{che2018recurrent}, Phased-LSTM \cite{neil2016phased}, and ODE-based models like LATENT-ODE and ODE-RNN \cite{chen2018neural}, our comparisons result contains mTAND, excluding results for the latter models.

As shown in \tableautorefname \ref{tab:main_result}, TimeCHEAT demonstrates competitive performance across the above three benchmark datasets, highlighting its effectiveness in typical time series classification tasks. In particular, for \textit{imbalanced} binary classification, TimeCHEAT outperforms the leading baselines on the P19 dataset and achieves competitive results on the P12 dataset, trailing the top performer by only $0.5\%$. But it stands out due to its lower time and space complexity compared to ViTST though achieves SOTA performance, converting 1D time series into 2D images potentially leading to significant space inefficiencies due to the introduction of extensive blank areas, especially problematic in ISMTS. On the more complex task of 8-class classification in the PAM dataset, TimeCHEAT surpasses existing methods, with a $0.7\%$ improvement in accuracy and a $0.9\%$ increase in precision. 

In almost all cases, our TimeCHEAT achieves consistent \textit{low standard deviation} indicating it is a reliable model. Its performance remains steady across varying data samples and initial conditions, suggesting a strong potential for generalizing well to new, unseen data. This stability and predictability in performance enhance the confidence in the model’s predictions, which is particularly crucial in sensitive areas such as medical diagnosis in clinical settings.

\subsection{Time Series Interpolation}
\paragraph{Datasets and experimental settings.}  
\textbf{PhysioNet} \cite{silva2012predicting} contains $37$ variables recorded during the first $48$ hours of ICU admission. For interpolation experiments, we utilize all $8,000$ instances, with a missing ratio of $78.0\%$.

The dataset is randomly split into $80\%$ for training and $20\%$ for testing, with $20\%$ of the training data set aside for validation. Performance is evaluated using MSE, where lower values indicate better performance.

\paragraph{Main Interpolation Results.}
For the interpolation task, we compare TimeCHEAT with RNN-VAE, L-ODE-RNN \cite{chen2018neural}, L-ODE-ODE \cite{rubanova2019latent}, and mTAND-full.

In this task, models are trained to predict or reconstruct values across the entire dataset based on a selected subset of observations. Experiments are conducted at varying observation levels, from $50\%$ to $90\%$ of observed points. During testing, models use the observed points to infer values at all time points within each test instance. We strictly follow the mTAND interpolation setup, where each column corresponds to a different setting. In each setting, a specific percentage of data is used to condition the model, which then predicts the remaining portion. Consequently, \textit{results across columns are not directly comparable}, but each reflects the performance of interpolation under its respective conditions. As illustrated in \tableautorefname \ref{table:phy_interp}, TimeCHEAT demonstrates great yet stable performance, highlighting its effectiveness in ISMTS interpolation.

\begin{table}[!t]
    \centering
    \scriptsize
    \tabcolsep=0.8em
    %\resizebox{0.788\textwidth}{!}{ 
    \begin{tabular}{l|cc|cc}
        \toprule
        & \multicolumn{2}{c|}{P19} & \multicolumn{2}{c}{P12}  \\ \cmidrule{2-5}
        \multirow{-2}{*}{Methods} & AUROC & AUPRC & AUROC & AUPRC  \\ \midrule
        w/o correlation  &\valstd{88.0}{3.1} & \valstd{54.4}{5.0} & \valstd{83.5}{0.9} & \valstd{46.5}{2.3}   \\
        w/ iTransformer   & \valstd{87.6}{2.4} & \valstd{54.7}{4.6}  &\valstd{79.9}{1.5} & \valstd{39.2}{3.2}  \\
        \makecell[l]{w/o correlation \\+ w/ iTransformer}  & \valstd{86.8}{2.7} & \valstd{54.2}{4.3} & \valstd{80.1}{1.7} & \valstd{39.4}{3.1} \\
        mTAND instead  & \valstd{87.4}{2.3} & \valstd{52.5}{3.4} & \valstd{84.3}{0.8} & \valstd{48.2}{1.0} \\
        \midrule
        \textbf{TimeCHEAT} &\valstdb{89.5}{1.9} &\valstdb{56.1}{4.6} &\valstdb{84.5}{0.7}  &\valstdb{48.2}{1.9} \\
        \bottomrule
    \end{tabular}
    %}
    \caption{Ablation studies on different strategies of TimeCHEAT in classification.  }
    \label{tab:ablation}
\end{table}

\subsection{Time Series Forecasting}
\paragraph{Datasets and Experimental Settings.}
(1) \textbf{USHCN} \cite{menne2015united} is a preprocessed dataset with measurements of $5$ variables from $1,280$ weather stations across the USA, featuring a missing ratio of $78.0\%$. (2) \textbf{MIMIC-III} \cite{johnson2016mimic} contains recorded observations of 96 variables at $30$-minute intervals, using data from the first 48 hours after ICU admission, with a missing ratio of $94.2\%$. (3) \textbf{MIMIC-IV} \cite{johnson2020mimic} is built upon the MIMIC-III database with a missing ratio up to $97.8\%$. It adopts a modular approach to data organization, highlighting data provenance and facilitating both individual and combined use of disparate data sources. (4) \textbf{Physionet12} \cite{silva2012predicting} includes medical records from $12,000$ ICU patients, capturing $37$ vital signs during the first $48$ hours of admission, with a missing ratio of $80.4\%$. The performance is evaluated using MSE.

\paragraph{Main Forecasting Results.}
We compare TimeCHEAT with various ISMTS forecasting models, including Grafiti \cite{grafiti2024Yalavarthi}, GRU-ODE-Bayes \cite{de2019gru}, Neural Flows \cite{bilovs2021neural}, CRU \cite{schirmer2022modeling}, NeuralODE-VAE \cite{chen2018neural}, GRUSimple, GRU-D, TLSTM \cite{baytas2017patient}, mTAND, and enhanced versions of Informer \cite{zhou2021informer}, Fedformer \cite{zhou2022fedformer}, DLinear, and NLinear \cite{zeng2023transformers}, referred to as Informer$+$, Fedformer$+$, DLinear$+$, and NLinear$+$.

Following the GraFITi setup, for the USHCN dataset, the model observes the first $3$ years and forecasts the next $3$ time steps. For the other datasets, the model observes the first $36$ hours and predicts the next $3$ time steps. As indicated in \tableautorefname{\ref{tab:fore}}, TimeCHEAT consistently shows competitive performance across all datasets, consistently ranking within the top two among baseline models. While GraFITi excels in scenarios where explicitly modeling the relationship between observation and prediction points is advantageous, TimeCHEAT remains highly competitive without relying on task-specific priors.

\subsection{Ablation Study}
We use two imbalance binary label datasets P12 and P19 in the classification task as an example to conduct the ablation study. We verify the necessity of three main designs: 1) learnable correlation between multiple channels in the embedding procedure, 2) channel-dependent embedding learning without special assumptions, and 3) channel-independent Transformer encoder. 

 As shown in \tableautorefname{\ref{tab:ablation}}, the full TimeCHEAT framework, which includes all original components (line 7), delivers the best performance. When local correlations between channels are removed (line 3), by replacing the channel encoding with a one-hot binary indicator vector (resulting in CI embedding learning), sparse sampling channels lack sufficient information and fail to aggregate crucial data from related channels for improved embedding. Discarding the CI encoder and switching to a vanilla Transformer (line 4) leads to a significant drop in classification accuracy, underscoring the effectiveness of the CI strategy in the encoding phase. Combining the above two changes results in a fully CI model (line 5), which yields nearly the worst accuracy among all tested conditions. Finally, replacing the local graph embedding with CI mTAND (line 5), while partially mitigating issues due to its assumptions about timestamp distances, still falls short of the best performance due to the limitations of the CI strategy.

\section{Conclusion}

In this paper, we introduce TimeCHEAT, a novel framework for ISMTS analysis. TimeCHEAT's innovative channel harmony strategy effectively balances individual channel processing with inter-channel interactions, improving ISMTS analysis performance. Our results show that combining the CD strategy locally with the CI strategy globally harnesses the strengths of both approaches, as well or better than a range of baseline and SOTA models. A key contribution of this work is to design the bipartite graphs with CD strategy locally to transform embedding learning into an edge weight prediction task, avoiding introducing inappropriate prior assumptions and enabling fixed-length embeddings for the encoder. Additionally, the CI strategy applied globally across patches allows the Transformer to learn individualized attention patterns for each channel, leading to superior representations for downstream tasks.

\section{Acknowledgements}
The authors wish to thank all the donors of the original datasets, and everyone who provided feedback on this work. This work is supported by the Key Program of NSFC under Grant No.62376126, Postgraduate Research \& Practice Innovation Program of Jiangsu Province under Grant No.KYCX21\_0225.

\bibliography{aaai25}

\begin{thebibliography}{60}
\providecommand{\natexlab}[1]{#1}

\bibitem[{Agarwal et~al.(2023)Agarwal, Sinha, Prasad, Clausel, Horsch, Constant, and Coubez}]{agarwal2023modelling}
Agarwal, R.; Sinha, A.; Prasad, D.~K.; Clausel, M.; Horsch, A.; Constant, M.; and Coubez, X. 2023.
\newblock Modelling Irregularly Sampled Time Series Without Imputation.
\newblock \emph{arXiv preprint arXiv:2309.08698}.

\bibitem[{Baytas et~al.(2017)Baytas, Xiao, Zhang, Wang, Jain, and Zhou}]{baytas2017patient}
Baytas, I.~M.; Xiao, C.; Zhang, X.; Wang, F.; Jain, A.~K.; and Zhou, J. 2017.
\newblock Patient subtyping via time-aware LSTM networks.
\newblock In \emph{ACM SIGKDD}, 65--74.

\bibitem[{Bilo{\v{s}} et~al.(2021)Bilo{\v{s}}, Sommer, Rangapuram, Januschowski, and G{\"u}nnemann}]{bilovs2021neural}
Bilo{\v{s}}, M.; Sommer, J.; Rangapuram, S.~S.; Januschowski, T.; and G{\"u}nnemann, S. 2021.
\newblock Neural flows: Efficient alternative to neural ODEs.
\newblock \emph{NeurIPS}, 34: 21325--21337.

\bibitem[{Camino, Hammerschmidt, and State(2019)}]{camino2019improving}
Camino, R.~D.; Hammerschmidt, C.~A.; and State, R. 2019.
\newblock Improving missing data imputation with deep generative models.
\newblock \emph{arXiv preprint arXiv:1902.10666}.

\bibitem[{Cao et~al.(2018)Cao, Wang, Li, Zhou, Li, and Li}]{cao2018brits}
Cao, W.; Wang, D.; Li, J.; Zhou, H.; Li, L.; and Li, Y. 2018.
\newblock Brits: Bidirectional recurrent imputation for time series.
\newblock \emph{NeurIPS}, 31.

\bibitem[{Che et~al.(2018)Che, Purushotham, Cho, Sontag, and Liu}]{che2018recurrent}
Che, Z.; Purushotham, S.; Cho, K.; Sontag, D.; and Liu, Y. 2018.
\newblock Recurrent neural networks for multivariate time series with missing values.
\newblock \emph{Scientific reports}, 8(1): 1--12.

\bibitem[{Chen et~al.(2024)Chen, Lenssen, Feng, Hu, Fey, Tassiulas, Leskovec, and Ying}]{chen2024similarity}
Chen, J.; Lenssen, J.~E.; Feng, A.; Hu, W.; Fey, M.; Tassiulas, L.; Leskovec, J.; and Ying, R. 2024.
\newblock From Similarity to Superiority: Channel Clustering for Time Series Forecasting.
\newblock \emph{arXiv preprint arXiv:2404.01340}.

\bibitem[{Chen et~al.(2018)Chen, Rubanova, Bettencourt, and Duvenaud}]{chen2018neural}
Chen, R.~T.; Rubanova, Y.; Bettencourt, J.; and Duvenaud, D.~K. 2018.
\newblock Neural ordinary differential equations.
\newblock \emph{NeurIPS}, 31.

\bibitem[{Chen et~al.(2022)Chen, Zhang, Zhao, Saunier, and Sun}]{chen2022nonstationary}
Chen, X.; Zhang, C.; Zhao, X.-L.; Saunier, N.; and Sun, L. 2022.
\newblock Nonstationary temporal matrix factorization for multivariate time series forecasting.
\newblock \emph{arXiv preprint arXiv:2203.10651}.

\bibitem[{Chowdhury et~al.(2023)Chowdhury, Li, Zhang, Hong, Gupta, and Shang}]{chowdhury2023primenet}
Chowdhury, R.~R.; Li, J.; Zhang, X.; Hong, D.; Gupta, R.~K.; and Shang, J. 2023.
\newblock Primenet: Pre-training for irregular multivariate time series.
\newblock In \emph{AAAI}, volume~37, 7184--7192.

\bibitem[{De~Brouwer et~al.(2019)De~Brouwer, Simm, Arany, and Moreau}]{de2019gru}
De~Brouwer, E.; Simm, J.; Arany, A.; and Moreau, Y. 2019.
\newblock GRU-ODE-Bayes: Continuous modeling of sporadically-observed time series.
\newblock \emph{NeurIPS}, 32.

\bibitem[{Du, C{\^o}t{\'e}, and Liu(2023)}]{du2023saits}
Du, W.; C{\^o}t{\'e}, D.; and Liu, Y. 2023.
\newblock Saits: Self-attention-based imputation for time series.
\newblock \emph{Expert Systems with Applications}, 219: 119619.

\bibitem[{Fan(2022)}]{fan2022dynamic}
Fan, J. 2022.
\newblock Dynamic Nonlinear Matrix Completion for Time-Varying Data Imputation.
\newblock In \emph{AAAI}.

\bibitem[{Goldberger et~al.(2000)Goldberger, Amaral, Glass, Hausdorff, Ivanov, Mark, Mietus, Moody, Peng, and Stanley}]{goldberger2000physiobank}
Goldberger, A.~L.; Amaral, L.~A.; Glass, L.; Hausdorff, J.~M.; Ivanov, P.~C.; Mark, R.~G.; Mietus, J.~E.; Moody, G.~B.; Peng, C.-K.; and Stanley, H.~E. 2000.
\newblock PhysioBank, PhysioToolkit, and PhysioNet: components of a new research resource for complex physiologic signals.
\newblock \emph{circulation}, 101(23): e215--e220.

\bibitem[{Han, Ye, and Zhan(2024)}]{han2024capacity}
Han, L.; Ye, H.-J.; and Zhan, D.-C. 2024.
\newblock The Capacity and Robustness Trade-off: Revisiting the Channel Independent Strategy for Multivariate Time Series Forecasting.
\newblock \emph{IEEE Transactions on Knowledge and Data Engineering}, (01): 1--14.

\bibitem[{He et~al.(2016)He, Zhang, Ren, and Sun}]{he2016deep}
He, K.; Zhang, X.; Ren, S.; and Sun, J. 2016.
\newblock Deep residual learning for image recognition.
\newblock In \emph{CVPR}, 770--778.

\bibitem[{Horn et~al.(2020)Horn, Moor, Bock, Rieck, and Borgwardt}]{horn2020set}
Horn, M.; Moor, M.; Bock, C.; Rieck, B.; and Borgwardt, K. 2020.
\newblock Set functions for time series.
\newblock In \emph{ICML}, 4353--4363. PMLR.

\bibitem[{Jhin et~al.(2022)Jhin, Lee, Jo, Kook, Jeon, Hyeong, Kim, and Park}]{jhin2022exit}
Jhin, S.~Y.; Lee, J.; Jo, M.; Kook, S.; Jeon, J.; Hyeong, J.; Kim, J.; and Park, N. 2022.
\newblock Exit: Extrapolation and interpolation-based neural controlled differential equations for time-series classification and forecasting.
\newblock In \emph{ACM Web Conference}, 3102--3112.

\bibitem[{Jin et~al.(2022)Jin, Zheng, Li, Chen, Yang, and Pan}]{jin2022multivariate}
Jin, M.; Zheng, Y.; Li, Y.-F.; Chen, S.; Yang, B.; and Pan, S. 2022.
\newblock Multivariate time series forecasting with dynamic graph neural odes.
\newblock \emph{IEEE Transactions on Knowledge and Data Engineering}.

\bibitem[{Johnson et~al.(2020)Johnson, Bulgarelli, Pollard, Horng, Celi, and Mark}]{johnson2020mimic}
Johnson, A.; Bulgarelli, L.; Pollard, T.; Horng, S.; Celi, L.~A.; and Mark, R. 2020.
\newblock Mimic-iv.
\newblock \emph{PhysioNet. Available online at: https://physionet. org/content/mimiciv/1.0/(accessed August 23, 2021)}, 49--55.

\bibitem[{Johnson et~al.(2016)Johnson, Pollard, Shen, Lehman, Feng, Ghassemi, Moody, Szolovits, Celi, and Mark}]{johnson2016mimic}
Johnson, A.; Pollard, T.~J.; Shen, L.; Lehman, L.-w.~H.; Feng, M.; Ghassemi, M.; Moody, B.; Szolovits, P.; Celi, L.~A.; and Mark, R.~G. 2016.
\newblock MIMIC-III, a freely accessible critical care database Sci.
\newblock \emph{Data}, 3(1): 1.

\bibitem[{Kenton and Toutanova(2019)}]{kenton2019bert}
Kenton, J. D. M.-W.~C.; and Toutanova, L.~K. 2019.
\newblock BERT: Pre-training of Deep Bidirectional Transformers for Language Understanding.
\newblock In \emph{Proceedings of NAACL-HLT}, 4171--4186.

\bibitem[{Kidger et~al.(2020)Kidger, Morrill, Foster, and Lyons}]{kidger2020neural}
Kidger, P.; Morrill, J.; Foster, J.; and Lyons, T. 2020.
\newblock Neural controlled differential equations for irregular time series.
\newblock \emph{NeurIPS}, 33: 6696--6707.

\bibitem[{Le~Morvan et~al.(2021)Le~Morvan, Josse, Scornet, and Varoquaux}]{le2021sa}
Le~Morvan, M.; Josse, J.; Scornet, E.; and Varoquaux, G. 2021.
\newblock What’s a good imputation to predict with missing values?
\newblock \emph{NeurIPS}, 34: 11530--11540.

\bibitem[{Li, Li, and Yan(2023)}]{li2023time}
Li, Z.; Li, S.; and Yan, X. 2023.
\newblock Time Series as Images: Vision Transformer for Irregularly Sampled Time Series.
\newblock In \emph{NeurIPS}.

\bibitem[{Li et~al.(2023)Li, Qi, Li, and Xu}]{li2023revisiting}
Li, Z.; Qi, S.; Li, Y.; and Xu, Z. 2023.
\newblock Revisiting long-term time series forecasting: An investigation on linear mapping.
\newblock \emph{arXiv preprint arXiv:2305.10721}.

\bibitem[{Liu et~al.(2021)Liu, Yu, Liao, Li, Lin, Liu, and Dustdar}]{liu2021pyraformer}
Liu, S.; Yu, H.; Liao, C.; Li, J.; Lin, W.; Liu, A.~X.; and Dustdar, S. 2021.
\newblock Pyraformer: Low-complexity pyramidal attention for long-range time series modeling and forecasting.
\newblock In \emph{ICLR}.

\bibitem[{Menne, Williams~Jr, and Vose(2015)}]{menne2015united}
Menne, M.~J.; Williams~Jr, C.; and Vose, R.~S. 2015.
\newblock United States historical climatology network daily temperature, precipitation, and snow data.
\newblock \emph{Carbon Dioxide Information Analysis Center, Oak Ridge National Laboratory, Oak Ridge, Tennessee}.

\bibitem[{Montero-Manso and Hyndman(2021)}]{montero2021principles}
Montero-Manso, P.; and Hyndman, R.~J. 2021.
\newblock Principles and algorithms for forecasting groups of time series: Locality and globality.
\newblock \emph{International Journal of Forecasting}, 37(4): 1632--1653.

\bibitem[{Murtagh and Heck(2012)}]{murtagh2012multivariate}
Murtagh, F.; and Heck, A. 2012.
\newblock \emph{Multivariate data analysis}, volume 131.
\newblock Springer Science \& Business Media.

\bibitem[{Neil, Pfeiffer, and Liu(2016)}]{neil2016phased}
Neil, D.; Pfeiffer, M.; and Liu, S.-C. 2016.
\newblock Phased lstm: Accelerating recurrent network training for long or event-based sequences.
\newblock \emph{NeurIPS}, 29.

\bibitem[{Nie et~al.(2023)Nie, Nguyen, Sinthong, and Kalagnanam}]{nie2023a}
Nie, Y.; Nguyen, N.~H.; Sinthong, P.; and Kalagnanam, J. 2023.
\newblock A Time Series is Worth 64 Words: Long-term Forecasting with Transformers.
\newblock In \emph{ICLR}.

\bibitem[{Reiss and Stricker(2012)}]{reiss2012introducing}
Reiss, A.; and Stricker, D. 2012.
\newblock Introducing a new benchmarked dataset for activity monitoring.
\newblock In \emph{16th international symposium on wearable computers}, 108--109. IEEE.

\bibitem[{Reyna et~al.(2020)Reyna, Josef, Jeter, Shashikumar, Westover, Nemati, Clifford, and Sharma}]{reyna2020early}
Reyna, M.~A.; Josef, C.~S.; Jeter, R.; Shashikumar, S.~P.; Westover, M.~B.; Nemati, S.; Clifford, G.~D.; and Sharma, A. 2020.
\newblock Early prediction of sepsis from clinical data: the PhysioNet/Computing in Cardiology Challenge 2019.
\newblock \emph{Critical care medicine}, 48(2): 210--217.

\bibitem[{Rubanova, Chen, and Duvenaud(2019)}]{rubanova2019latent}
Rubanova, Y.; Chen, R.~T.; and Duvenaud, D.~K. 2019.
\newblock Latent ordinary differential equations for irregularly-sampled time series.
\newblock \emph{NeurIPS}, 32.

\bibitem[{Schirmer et~al.(2022)Schirmer, Eltayeb, Lessmann, and Rudolph}]{schirmer2022modeling}
Schirmer, M.; Eltayeb, M.; Lessmann, S.; and Rudolph, M. 2022.
\newblock Modeling irregular time series with continuous recurrent units.
\newblock In \emph{ICML}, 19388--19405. PMLR.

\bibitem[{Scholz et~al.(2022)Scholz, Born, Duong-Trung, Cruz-Bournazou, and Schmidt-Thieme}]{scholz2022latent}
Scholz, R.; Born, S.; Duong-Trung, N.; Cruz-Bournazou, M.~N.; and Schmidt-Thieme, L. 2022.
\newblock Latent Linear ODEs with Neural Kalman Filtering for Irregular Time Series Forecasting.
\newblock \emph{NeurIPS}.

\bibitem[{Schulz and Stattegger(1997)}]{schulz1997spectrum}
Schulz, M.; and Stattegger, K. 1997.
\newblock SPECTRUM: Spectral analysis of unevenly spaced paleoclimatic time series.
\newblock \emph{Computers \& Geosciences}, 23(9): 929--945.

\bibitem[{Shukla and Marlin(2018)}]{shukla2018interpolation}
Shukla, S.~N.; and Marlin, B. 2018.
\newblock Interpolation-Prediction Networks for Irregularly Sampled Time Series.
\newblock In \emph{ICLR}.

\bibitem[{Shukla and Marlin(2021)}]{shukla2021multitime}
Shukla, S.~N.; and Marlin, B. 2021.
\newblock Multi-Time Attention Networks for Irregularly Sampled Time Series.
\newblock In \emph{ICLR}.

\bibitem[{Shukla and Marlin(2022)}]{shukla2022heteroscedastic}
Shukla, S.~N.; and Marlin, B. 2022.
\newblock Heteroscedastic Temporal Variational Autoencoder For Irregularly Sampled Time Series.
\newblock In \emph{ICLR}.

\bibitem[{Silva et~al.(2012)Silva, Moody, Scott, Celi, and Mark}]{silva2012predicting}
Silva, I.; Moody, G.; Scott, D.~J.; Celi, L.~A.; and Mark, R.~G. 2012.
\newblock Predicting in-hospital mortality of icu patients: The physionet/computing in cardiology challenge 2012.
\newblock In \emph{2012 Computing in Cardiology}, 245--248. IEEE.

\bibitem[{Sun et~al.(2024)Sun, Li, Song, Cai, Zhang, and Hong}]{sun2024time}
Sun, C.; Li, H.; Song, M.; Cai, D.; Zhang, B.; and Hong, S. 2024.
\newblock Time pattern reconstruction for classification of irregularly sampled time series.
\newblock \emph{Pattern Recognition}, 147: 110075.

\bibitem[{Tan et~al.(2023)Tan, Li, Gao, Guan, Wang, Liu, Wu, and Li}]{tan2023openstl}
Tan, C.; Li, S.; Gao, Z.; Guan, W.; Wang, Z.; Liu, Z.; Wu, L.; and Li, S.~Z. 2023.
\newblock Openstl: A comprehensive benchmark of spatio-temporal predictive learning.
\newblock \emph{NeurIPS}, 36: 69819--69831.

\bibitem[{Tang et~al.(2020)Tang, Yao, Sun, Aggarwal, Mitra, and Wang}]{tang2020joint}
Tang, X.; Yao, H.; Sun, Y.; Aggarwal, C.; Mitra, P.; and Wang, S. 2020.
\newblock Joint modeling of local and global temporal dynamics for multivariate time series forecasting with missing values.
\newblock In \emph{AAAI}, volume~34, 5956--5963.

\bibitem[{Tashiro et~al.(2021)Tashiro, Song, Song, and Ermon}]{tashiro2021csdi}
Tashiro, Y.; Song, J.; Song, Y.; and Ermon, S. 2021.
\newblock CSDI: Conditional Score-based Diffusion Models for Probabilistic Time Series Imputation.
\newblock \emph{NeurIPS}, 34.

\bibitem[{Vaswani et~al.(2017)Vaswani, Shazeer, Parmar, Uszkoreit, Jones, Gomez, Kaiser, and Polosukhin}]{vaswani2017attention}
Vaswani, A.; Shazeer, N.; Parmar, N.; Uszkoreit, J.; Jones, L.; Gomez, A.~N.; Kaiser, {\L}.; and Polosukhin, I. 2017.
\newblock Attention is all you need.
\newblock \emph{NeurIPS}, 30.

\bibitem[{Veli{\v{c}}kovi{\'c} et~al.(2017)Veli{\v{c}}kovi{\'c}, Cucurull, Casanova, Romero, Lio, and Bengio}]{velivckovic2017graph}
Veli{\v{c}}kovi{\'c}, P.; Cucurull, G.; Casanova, A.; Romero, A.; Lio, P.; and Bengio, Y. 2017.
\newblock Graph attention networks.
\newblock \emph{arXiv preprint arXiv:1710.10903}.

\bibitem[{Wang et~al.(2024)Wang, Du, Cao, Zhang, Wang, Liang, and Wen}]{wang2024deep}
Wang, J.; Du, W.; Cao, W.; Zhang, K.; Wang, W.; Liang, Y.; and Wen, Q. 2024.
\newblock Deep Learning for Multivariate Time Series Imputation: A Survey.
\newblock \emph{arXiv preprint arXiv:2402.04059}.

\bibitem[{Wu et~al.(2021{\natexlab{a}})Wu, Xu, Wang, and Long}]{wu2021autoformer}
Wu, H.; Xu, J.; Wang, J.; and Long, M. 2021{\natexlab{a}}.
\newblock Autoformer: Decomposition transformers with auto-correlation for long-term series forecasting.
\newblock \emph{NeurIPS}, 34: 22419--22430.

\bibitem[{Wu et~al.(2021{\natexlab{b}})Wu, Ni, Cheng, Zong, Song, Chen, Liu, Zhang, Chen, and Davidson}]{wu2021dynamic}
Wu, Y.; Ni, J.; Cheng, W.; Zong, B.; Song, D.; Chen, Z.; Liu, Y.; Zhang, X.; Chen, H.; and Davidson, S.~B. 2021{\natexlab{b}}.
\newblock Dynamic gaussian mixture based deep generative model for robust forecasting on sparse multivariate time series.
\newblock In \emph{AAAI}, volume~35, 651--659.

\bibitem[{Wu et~al.(2020)Wu, Pan, Long, Jiang, Chang, and Zhang}]{wu2020connecting}
Wu, Z.; Pan, S.; Long, G.; Jiang, J.; Chang, X.; and Zhang, C. 2020.
\newblock Connecting the dots: Multivariate time series forecasting with graph neural networks.
\newblock In \emph{ACM SIGKDD}, 753--763.

\bibitem[{Yalavarthi et~al.(2024)Yalavarthi, Madhusudhanan, Scholz, Ahmed, Burchert, Jawed, Born, and Schmidt-Thieme}]{grafiti2024Yalavarthi}
Yalavarthi, V.~K.; Madhusudhanan, K.; Scholz, R.; Ahmed, N.; Burchert, J.; Jawed, S.; Born, S.; and Schmidt-Thieme, L. 2024.
\newblock GraFITi: Graphs for Forecasting Irregularly Sampled Time Series.
\newblock In \emph{AAAI}, 16255--16263.

\bibitem[{Yoon, Jordon, and Schaar(2018)}]{yoon2018gain}
Yoon, J.; Jordon, J.; and Schaar, M. 2018.
\newblock Gain: Missing data imputation using generative adversarial nets.
\newblock In \emph{ICML}, 5689--5698. PMLR.

\bibitem[{You et~al.(2020)You, Ma, Ding, Kochenderfer, and Leskovec}]{you2020handling}
You, J.; Ma, X.; Ding, Y.; Kochenderfer, M.~J.; and Leskovec, J. 2020.
\newblock Handling missing data with graph representation learning.
\newblock \emph{NeurIPS}, 33: 19075--19087.

\bibitem[{Zeng et~al.(2023)Zeng, Chen, Zhang, and Xu}]{zeng2023transformers}
Zeng, A.; Chen, M.; Zhang, L.; and Xu, Q. 2023.
\newblock Are transformers effective for time series forecasting?
\newblock In \emph{AAAI}, volume~37, 11121--11128.

\bibitem[{Zhang et~al.(2021{\natexlab{a}})Zhang, Zeman, Tsiligkaridis, and Zitnik}]{zhang2021graph}
Zhang, X.; Zeman, M.; Tsiligkaridis, T.; and Zitnik, M. 2021{\natexlab{a}}.
\newblock Graph-Guided Network for Irregularly Sampled Multivariate Time Series.
\newblock In \emph{ICLR}.

\bibitem[{Zhang et~al.(2021{\natexlab{b}})Zhang, Zhang, Jiang, and Zhou}]{zhang2021life}
Zhang, Z.-Y.; Zhang, S.-Q.; Jiang, Y.; and Zhou, Z.-H. 2021{\natexlab{b}}.
\newblock LIFE: Learning individual features for multivariate time series prediction with missing values.
\newblock In \emph{ICDM}, 1511--1516. IEEE.

\bibitem[{Zhou et~al.(2021)Zhou, Zhang, Peng, Zhang, Li, Xiong, and Zhang}]{zhou2021informer}
Zhou, H.; Zhang, S.; Peng, J.; Zhang, S.; Li, J.; Xiong, H.; and Zhang, W. 2021.
\newblock Informer: Beyond efficient transformer for long sequence time-series forecasting.
\newblock In \emph{AAAI}, volume~35, 11106--11115.

\bibitem[{Zhou et~al.(2022)Zhou, Ma, Wen, Wang, Sun, and Jin}]{zhou2022fedformer}
Zhou, T.; Ma, Z.; Wen, Q.; Wang, X.; Sun, L.; and Jin, R. 2022.
\newblock Fedformer: Frequency enhanced decomposed transformer for long-term series forecasting.
\newblock In \emph{ICML}, 27268--27286. PMLR.

\end{thebibliography}

\end{document}